\documentclass[10pt,letterpaper,twocolumn]{article}
\usepackage[T1]{fontenc}
\usepackage[utf8]{inputenc}
\usepackage{mathptmx}
\usepackage[margin=0.6in,columnsep=0.25in]{geometry}
\usepackage{amsmath,amssymb}
\usepackage{graphicx}
\usepackage{array,booktabs,tabularx}
\usepackage{float}
\usepackage{caption}
\usepackage{enumitem}
\usepackage{pgfplots}
\usepackage[hidelinks]{hyperref}
\newcommand{\papercite}[1]{\hyperlink{paper-ref-#1}{[#1]}}
\pgfplotsset{compat=1.18}
\setlist[itemize]{leftmargin=1.2em,itemsep=2pt,topsep=3pt,parsep=0pt}
\newcolumntype{Y}{>{\raggedright\arraybackslash}X}
\definecolor{policygray}{RGB}{139,140,131}
\definecolor{policygreen}{RGB}{82,194,159}
\definecolor{policyblue}{RGB}{47,90,158}
\pgfplotsset{paperplot/.style={
  width=\columnwidth,height=1.85in,
  scale only axis=false,
  tick label style={font=\scriptsize},label style={font=\scriptsize},
  title style={font=\scriptsize\bfseries,align=center},
  axis line style={black!70},tick style={black!70},
  ymajorgrids=true,grid style={black!12},
  enlarge x limits=0.23,
  nodes near coords style={font=\scriptsize\bfseries},
}}
\title{Continuous Risk-Based Scheduling for Food Safety Inspection:\\
Integrating Historical Decay with Event-Driven Consumer Review Signals}
\author{Akriti Bhat\textsuperscript{1}, Wayne Zhu\textsuperscript{2}, Ishan Gupta\textsuperscript{3}}
\date{}
\begin{document}
\maketitle
\raggedbottom

\begin{abstract}
Batch risk-based inspection scheduling struggles to detect a critical failure class: food facilities with recent clean inspections that subsequently develop severe violations. Traditional systems recompute risk on fixed planning cycles using static historical data, leaving blind spots between inspection cycles. To address this, we introduce the Continuous Inspection Risk (CIR) system, an event-driven scheduling pipeline that re-ranks the inspection queue dynamically as new data arrives. CIR combines a decay-weighted historical score with a stream of real-time consumer reviews triaged by a language-model classifier. A key structural contribution is our temporal partition rule, which strictly excludes reviews predating the most recent inspection to prevent superseded signals from skewing current risk. Evaluating CIR on 9,505 routine inspections in Santa Clara County, California (2024--2026), we show that history-based scoring alone reduces mean time-to-detection of major violations by 69 days (20.9\%) under a 30\% inspection budget, but fails to anticipate severe vermin-related closures (0/8 detected). Incorporating post-inspection review signals drastically improves performance, predicting 7 of 8 vermin closures with an average lead time of 117 days-firing on the exact day of review publication in four instances. Introducing a review corroboration rule improves precision from 0.70 to 0.78 without loss of recall. Finally, we analyze critical implementation challenges, documenting a data-hygiene pitfall (stale review leakage across inspection boundaries) and an evaluation limitation involving unobserved facility self-remediation. This work provides the first case-level evaluation of event-driven inspection triggering using open data and provides a reproducible benchmark for review-augmented public health systems.
\end{abstract}


\section{Introduction}

When the City of Chicago deployed the first machine-learned food inspection forecasting system in 2015, the promise was straightforward: rank restaurants by predicted risk, send inspectors to the highest-risk venues first, and detect critical violations weeks earlier than traditional schedules \papercite{1}. The pilot reported success, and risk-based scheduling quickly entered the standard toolkit of data-driven municipal governance. A decade later, however, an independent retrospective complicated this narrative: the model's most influential feature was the cluster of the inspector who had conducted the restaurant's most recent inspection (a proxy that could reflect differences in inspector strictness or in the underlying risk of establishments assigned to them, not a causal effect of the inspector on later conditions), and the causal link between model predictions and reported health gains could not be established \papercite{2}. Nevertheless, the underlying architecture expanded nationwide, carrying with it a foundational assumption worth examining.

Every system in this lineage assumes that facility risk changes slowly-adequately captured by recomputing scores at fixed planning cycles and holding them static until the next run. Under this assumption, a facility that passes an inspection cleanly in July but develops a severe rodent infestation in September is structurally invisible to the system until the next batch run-regardless of how many patrons dine there in the interim. Foodborne illness affects tens of millions of Americans annually, with commercial food establishments accounting for the majority of outbreaks \papercite{3}. The facilities trapped in this operational blindspot-clean paperwork, deteriorating kitchen-cannot be caught by history-scored models under any parameterization, because the only signal those models process is the very record that appears clean.

We study an event-driven alternative that eliminates the waiting period of fixed planning cycles. Our approach maintains a continuously updated risk score per facility, combining a slow signal (decay-weighted inspection history) with a fast signal (consumer reviews classified for food-safety hazards in real time), re-ranking the inspection queue the moment qualifying evidence arrives. Unlike illness-related search surveillance, which inherently triggers only after a patron falls ill, consumer review text can capture visible hazards-such as pest sightings or poor hygiene-before harm occurs. Our primary empirical question is narrow and falsifiable: \textit{In facilities whose most recent inspection was clean but were later closed for vermin violations, can an event-driven signal flag them prior to closure when historical data alone cannot prove?}

\subsection*{Contributions}

Empirical Backtest of History-Based Risk: We demonstrate on 9,505 Santa Clara County inspections (snapshot June 2026) that a single-parameter decay score detects major violations 69 days faster (a 20.9\% improvement) than chronological scheduling under a 30\% inspection budget (Section 5).

Proof of Structural Blindspots: We present a case-level evaluation proving that history-based scoring has an absolute structural floor, failing to anticipate any documented vermin-related closures (0 of 8) prior to occurrence, regardless of parameterization (Section 6.1).

Data-Hygiene Hazard Analysis: We document a critical data-hygiene pitfall : pre-window review leakage across inspection boundaries, that artificially inflated one of our own lead times by 100 days before correction, offering a warning taxonomy for systems joining archival review streams with inspection records (Section 6.2).

Audited Live-Review Trigger Performance: We evaluate an event-driven review trigger that achieves 7/8 recall with a 117-day mean lead time (firing same-day in 4 of 7 instances). Introducing a corroboration rule improves precision from 0.70 to 0.78 with zero recall cost on this case set (Section 7).

Evaluation Limits \& Remediation Dynamics: We provide an explicit account of an inherent evaluation constraint: why closure-based ground truth cannot distinguish unobserved false positives from quiet, successful facility remediation, directly connecting our findings to the Chicago retrospective literature (Section 8).

\section{Related Work}

Batch risk scoring from inspection history: Chicago's forecasting model \papercite{1} ranks facilities using prior inspection outcomes, 311 complaint records, and establishment attributes. A pilot deployment reported finding critical violations weeks earlier than the incumbent schedule. However, an independent retrospective by Kannan et al. \papercite{2} re-examined the model and raised serious reservations: the single most influential feature was the cluster of the inspector who had conducted the facility's most recent inspection, meaning predicted risk changed substantially according to who had last inspected it (a factor outside the facility's control), the feature set was too coarse to anticipate several failure classes, and the causal link between the model and reported gains could not be established due to unaddressed confounds such as inspector turnover. We make no comparative performance claim against this system---our data, jurisdiction, and baseline construction all differ---and cite its retrospective chiefly because our own evaluation inherits a version of the same ground-truth ambiguity (Section 8).

Real-time triggering from private telemetry: The nearest neighbor to our work is FINDER \papercite{7}, a production system developed by Google and Harvard deployed across health departments in Las Vegas and Chicago. FINDER infers likely outbreak sources from de-identified, aggregated web search and location data---computing the fraction of venue visitors who subsequently searched for illness-related terms---and prompts inspectors to visit specific restaurants off-cycle. Across both cities, restaurants flagged by FINDER were found unsafe 52.3\% of the time, compared to 22.7\% for routine inspections. While FINDER establishes that continuous off-cycle triggering works at scale, our contribution is deliberately positioned downstream of that result:

\begin{itemize}
\item Data Accessibility: FINDER's input stream exists exclusively inside a proprietary search-and-location provider. Our system consumes only publicly posted review text, which any municipal agency can collect without corporate partnerships.

\item Pre-Illness Signal Potential: An illness-inference signal fires only after symptoms emerge, whereas consumer reviews can describe pre-illness hazards (e.g., visible pests)---a structural distinction we note but do not quantitatively isolate.

\item Scope \& Pitfalls: While our offline evaluation (8 documented closures) is smaller than FINDER's deployed randomized trial, we test whether continuous triggering survives the transition to public data and identify a unique data-hygiene pitfall (Section 6.2) inherent to archival review streams.

\end{itemize}
Review text as a predictor of inspection outcomes: Multiple studies establish that review text and star ratings correlate with official inspection results, spanning keyword classifiers on Yelp \papercite{4} to multi-city social media surveillance \papercite{5}. However, these are population-level, batch analyses that demonstrate macro correlation rather than operational scheduling logic. They do not evaluate whether a facility's queue position should change the day a specific review appears, nor do they combine review streams with decay-weighted inspection histories. Evaluating that dynamic combination on a case-by-case basis against documented closures is the exact gap we address.

Language models on informal text: Zero- and few-shot classification of noisy, user-generated content with large language models is now well established, including in health-adjacent domains \papercite{6}. We leverage this capability directly, employing an off-the-shelf model as a zero-shot triage classifier rather than training a task-specific model, perfectly matching the small-label regime of our domain.

\section{System Architecture: Three Evidence Layers}

The system maintains a live risk score for each facility and recomputes it dynamically whenever new evidence arrives rather than on a fixed calendar cycle. Evidence is strictly partitioned by its temporal relationship to the facility's most recent completed inspection:

\begin{itemize}
\item Layer 1 (Historical Decay): A parametric transform of the most recent inspection score, discounted exponentially as that inspection ages. This represents the slow signal baseline used by prior batch systems.

\item Layer 2 (Pre-Window Review History): Reviews dated \textit{before} the most recent inspection. Because the conditions described in these reviews were already evaluated during that inspection, treating them as live evidence would double-count past risk. The system retains Layer 2 records for auditability but structurally excludes them from score generation.

\item Layer 3 (Live Review Signal): Reviews posted \textit{after} the most recent inspection, classified on arrival by the triage model. This is the fast signal---only Layer 3 evidence can trigger off-cycle queue promotions.

\end{itemize}
The Continuous Inspection Risk (CIR) score on day $t$ following the most recent inspection is defined as:

\[
\operatorname{CIR}(t) = \alpha\,(100-S_{\mathrm{last}})\,e^{-\lambda t} + B_{\mathrm{live}}(t)
\]

Here, $t$ is the number of days since the facility's most recent inspection; $S_{\mathrm{last}}$ is the most recent inspection score; $\alpha$ scales the historical inspection-risk component; and $\lambda$ controls how quickly that component decays over time. The live review boost $B_{\mathrm{live}}(t)$ sums $\beta \times \text{confidence} \times \text{severity}$ across all Layer 3 reviews posted on or before day $t$, where $\beta$ is the live-review weight. Parameter values are reported in Appendix A. When $\operatorname{CIR}(t)$ reaches the scheduling threshold, the facility is promoted immediately.

\subsection*{Corroboration Rule}

The corroboration rule limits the influence of Layer 3 when only one qualifying post-inspection safety review is available for a facility; it does not penalize facilities merely for having few reviews overall. A single qualifying review is still included, but to prevent one anonymous complaint from overriding a clean official record, it is discounted to 0.3 \ensuremath{\times} its nominal weight by default. Here, a single review means one qualifying Layer 3 review record, regardless of how many hazards that review mentions. However, we enforce two domain-specific exceptions:

\begin{itemize}
\item Prior Risk Exception: If the facility's most recent inspection score was already below 90, the review is applied at full weight (1.0 \ensuremath{\times}), as the baseline record was not clean.

\item High-Severity Exception: If the review is both highly confident (\textit{\ensuremath{\geq} }0.95)  and severe (\textit{\ensuremath{\geq} }8, e.g. a specific illness or contamination report), it is applied at 0.6\textit{\ensuremath{\times} } weight, even in isolation.

\end{itemize}
Two or more separate qualifying Layer 3 reviews are treated as corroborating evidence and are always evaluated at full weight (1.0\ensuremath{\times}). Both the baseline corroboration rule and its severity exception were validated against a non-vermin control group to measure induced false positives (Section 7.1), rather than overfitted to the closure subset.

\section{Data}

\subsection{Inspection Records}

We utilize public inspection records from the Santa Clara County Department of Environmental Health covering July 2024 through June 2026. After merging multi-year county records with official violation tables, the resulting dataset contains 9,505 routine inspections with valid scores. Records are keyed by facility permit; per-facility temporal features (e.g., days since prior inspection, historical major-violation rates, prior inspection scores) are derived directly from this merged dataset.

\subsection{Review Corpus}

To evaluate off-cycle triggering, we collected Google Maps and Yelp reviews spanning the months leading up to documented events across three facility categories:

\begin{itemize}
\item Vermin Closures: 8 restaurants with documented vermin-related emergency closures.

\item High-Scoring Baseline Controls: 6 high-scoring comparison restaurants (score \ensuremath{\geq} 95).

\item Non-Pest Closures: 3 control facilities closed for non-pest operational issues (e.g., water service failures).

\end{itemize}
Crucially, no content-based pre-filtering was applied during collection; every review published within the temporal window was processed by the classifier to avoid selection bias toward alarming text. Each review was subsequently assigned to Layer 2 (pre-inspection) or Layer 3 (post-inspection) based on its publication date relative to the facility's most recent official inspection (Section 6.2).

\subsection{Triage Classifier Check}

The triage classifier labels each review as either SAFETY RISK or NOT RELEVANT, outputting a confidence score alongside the specific text snippet that triggered the classification. On a 30-review hand-labeled validation set, the classifier achieved 90\% accuracy, demonstrating strong sensitivity to food safety hazards (Recall: 0.94, Precision: 0.89).

\begin{table}[H]
\centering
\footnotesize
\begin{tabularx}{\columnwidth}{@{}YYY@{}}
\toprule
 & \textbf{\textbf{Predicted SAFETY RISK}} & \textbf{\textbf{Predicted NOT RELEVANT}} \\
\midrule
Actual SAFETY RISK & 16 (TP) & 1 (FN) \\
Actual NOT RELEVANT & 2 (FP) & 11 (TN) \\
\bottomrule
\end{tabularx}
\caption*{\textit{Table 1. Confusion matrix for the triage classifier on the 30-review hand-labeled validation set.}}
\end{table}

\begin{itemize}
\item Error Analysis: The single false negative cited a vague "unpleasant smell" without describing specific pests, symptoms, or hygiene breaches. The two false positives described mild indigestion without establishing a clear link to the meal.

\item Confidence Calibration: Ambiguous reviews yielded visibly lower model confidence (0.60--0.70) compared to explicit reports naming pests or acute illness (0.85--0.99). We treat this 30-item evaluation as an initial validation check rather than a population-wide estimate; Section 9 outlines planned large-scale evaluation protocols.

\end{itemize}
\section{Layer 1 Alone: Efficiency at Scale}

Before introducing review streams, we establish the performance of the historical decay layer (Magent) alone using the Santa Clara County dataset of 9,505 routine inspections from July 2024 through June 2026. We compare three scheduling policies on 7,953 held-out routine inspections from July 2025 onward. Figures 1 and 2 report major-violation catch rates and mean times to detection, respectively.

\begin{itemize}
\item \textit{M}\textsubscript{0} (Chronological Baseline): Inspect facilities strictly in order of elapsed time since their last inspection---representing the standard status quo practice.

\item \textit{M}\textsubscript{1b} (Score Threshold Heuristic): Prioritize any facility whose most recent inspection score fell below 90---a common administrative heuristic for targeting risky venues.

\item \textit{M}\textsubscript{agent} (Layer 1 Historical Decay Model): Rank facilities continuously by their exponential decay score (CIR\textsubscript{Layer 1}) and deploy inspectors to the top 30\% highest-risk venues under a fixed capacity constraint.

\end{itemize}
\begin{figure}[H]
\centering
\begin{tikzpicture}
\begin{axis}[paperplot,ybar,bar width=28pt,ymin=0,ymax=110,
 title={Major Violation Catch Rate (\%)},ylabel={\%},
 xtick={0,1,2},xticklabels={\shortstack{M0: Chronological\\(100\% budget)},\shortstack{M1b: Score Threshold\\(36\% budget)},\shortstack{Magent: Ours\\(30\% budget)}},
 xticklabel style={font=\tiny,align=center},
 nodes near coords={\pgfmathprintnumber[fixed,precision=1,zerofill]{\pgfplotspointmeta}\%}]
\addplot[fill=policygray,draw=none,bar shift=0pt] coordinates {(0,89.7)};
\addplot[fill=policygreen,draw=none,bar shift=0pt] coordinates {(1,65.0)};
\addplot[fill=policyblue,draw=none,bar shift=0pt] coordinates {(2,49.0)};
\end{axis}
\end{tikzpicture}
\caption{Major-violation catch rates under three inspection scheduling policies. On 7,953 held-out routine inspections from Santa Clara County, chronological scheduling (M0) achieves an 89.7\% catch rate at a 100\% inspection budget, the score-threshold heuristic (M1b; prior score < 90) achieves 65.0\% at a 36\% budget, and the Layer 1 historical decay model (Magent) achieves 49.0\% at a 30\% budget.}
\label{fig:1}
\end{figure}
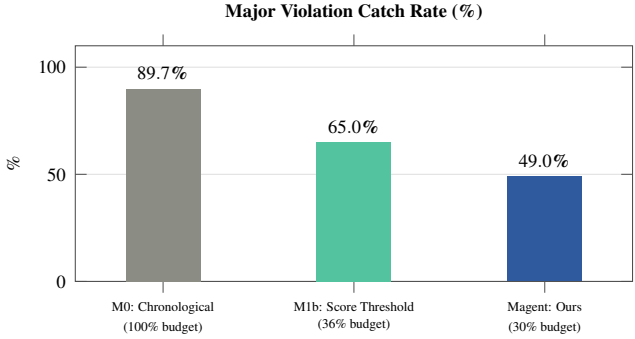

\begin{figure}[H]
\centering
\begin{tikzpicture}
\begin{axis}[paperplot,ybar,bar width=28pt,ymin=0,ymax=380,
 title={Mean Days to Detection\\(lower = faster intervention)},ylabel={Days},
 xtick={0,1,2},xticklabels={\shortstack{M0: Chronological\\(100\% budget)},\shortstack{M1b: Score Threshold\\(36\% budget)},\shortstack{Magent: Ours\\(30\% budget)}},
 xticklabel style={font=\tiny,align=center},
 nodes near coords={\pgfmathprintnumber[fixed,precision=1,zerofill]{\pgfplotspointmeta}d}]
\addplot[fill=policygray,draw=none,bar shift=0pt] coordinates {(0,329.7)};
\addplot[fill=policygreen,draw=none,bar shift=0pt] coordinates {(1,312.3)};
\addplot[fill=policyblue,draw=none,bar shift=0pt] coordinates {(2,260.7)};
\draw[red,thick,->] (axis cs:0,329.7) -- (axis cs:2,260.7);
\node[red,font=\tiny\bfseries,align=center,anchor=south] at (axis cs:1.85,300) {$-69$ days\\($-20.9\%$)};
\end{axis}
\end{tikzpicture}
\caption{Mean time to major-violation detection under three inspection scheduling policies. On the same 7,953 held-out routine inspections, mean time to detection is 329.7 days for chronological scheduling (M0), 312.3 days for the score-threshold heuristic (M1b), and 260.7 days for the Layer 1 historical decay model (Magent). Magent reduces mean time to detection by 69 days (20.9\%) relative to M0, with inspection budgets of 30\% and 100\%, respectively.}
\label{fig:2}
\end{figure}
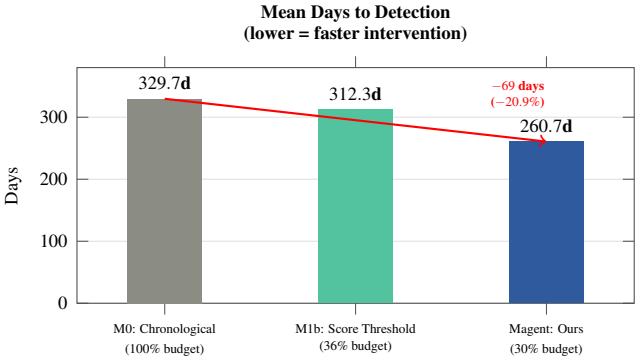

\subsection{Ablation}

Decomposing the decay score shows the exponential-decay term carries essentially all of the gain; a term for the facility's historical major-violation rate adds nothing measurable and is insensitive to its weight across the swept range. For deployment we would drop it and keep the one-parameter decay form. In Table 1, budget denotes the percentage of held-out facilities selected for inspection by each scheduling policy.

\begin{table}[H]
\centering
\footnotesize
\begin{tabularx}{\columnwidth}{@{}>{\hsize=1.7\hsize\linewidth=\hsize}Y>{\hsize=.6\hsize\linewidth=\hsize}Y>{\hsize=.8\hsize\linewidth=\hsize}Y>{\hsize=.9\hsize\linewidth=\hsize}Y@{}}
\toprule
\textbf{\textbf{Scheduler}} & \textbf{\textbf{Budget}} & \textbf{\textbf{Catch Rate}} & \textbf{\textbf{Mean Days to Detection}} \\
\midrule
\textit{M}\textsubscript{0} (chronological) & 100\% & 89.7\% & 329.7 \\
\textit{M}\textsubscript{1b} (score < 90) & 36\% & 65.0\% & 312.3 \\
\textit{M}\textsubscript{agent} (Layer 1 only) & 30\% & 49.0\% & 260.7 \\
\bottomrule
\end{tabularx}
\caption*{\textit{Table 1. Layer 1 backtest on 7,953 held-out inspections. }}
\end{table}

\subsection{Additional Baselines}

Two further baselines bound the result: random selection of 30\% of facilities (a floor - it carries no information and performs accordingly) and a reviews-only trigger that schedules any facility with at least one live SAFETY RISK review regardless of score, which reappears as the naive configuration in Section 7.1.

\section{Where Layer 1 Fails, and Why Layer 2 Must Stay Out}

We now narrow deliberately from 9,505 inspections to eight documented vermin closures - a case study, not a population sample - to ask two questions: why the decay layer misses these specific facilities, and what happens if pre-window reviews are allowed to leak into the live signal.

\subsection{Zero of Eight from History Alone}

\begin{table}[H]
\centering
\footnotesize
\begin{tabularx}{\columnwidth}{@{}>{\hsize=1.7\hsize\linewidth=\hsize}Y>{\hsize=.6\hsize\linewidth=\hsize}Y>{\hsize=.8\hsize\linewidth=\hsize}Y>{\hsize=.9\hsize\linewidth=\hsize}Y@{}}
\toprule
\textbf{\textbf{Restaurant}} & \textbf{\textbf{Last Score}} & \textbf{\textbf{Layer 1 Flags It?}} & \textbf{\textbf{Closure Cause}} \\
\midrule
Red Robin & 91 & No & Vermin \\
Pacific Catch & 82 & No & Vermin \\
Benihana & 88 & No & Vermin \\
Round Table Pizza & 88 & No & Vermin \\
Wienerschnitzel & 82 & No & Vermin \\
Black Bear Diner & 82 & No\textsuperscript{1} & Vermin \\
The Counter & 72 & No\textsuperscript{1} & Vermin \\
Bikaner Sweet & 75 & No\textsuperscript{1} & Vermin \\
\bottomrule
\end{tabularx}
\caption*{\textit{Table 2. Layer 1 outcome per case. }\textit{\textsuperscript{1 }}\textit{Even the three facilities with visibly sliding scores never reach the scheduling threshold on decay alone within the available window: at these score levels the decay term tops out well below it. Total: 0 of 8. }}
\end{table}

The result is structural rather than a tuning artifact. Five of the eight had final scores of 82--91 - records an inspector would read as acceptable - and for the remaining three the decay term is arithmetically incapable of reaching the threshold from scores in the 70s within the time available. No reweighting of \textit{\ensuremath{\alpha} }or \textit{\ensuremath{\lambda} }changes this without also flooding the county's whole facility population into the queue.

\subsection{The Stale-Review Leak: A Correction We Almost Missed}

An early version of our simulation compared each review's date against the wrong reference point, allowing some Layer 2 reviews to act as if they were live. Three of the eight restaurants were affected. The most visible symptom was a reported lead time of 309 days for one facility - which, on inspection, meant the system had been credited with reacting to a review posted eight months before the monitoring window even opened.

\begin{figure}[H]
\centering
\begin{tikzpicture}
\begin{axis}[paperplot,ybar,bar width=32pt,ymin=0,ymax=9,
 title={Recall by Evidence Configuration on the 8-Case Closure Set},
 ylabel={Vermin closures anticipated (of 8)},ytick={0,1,...,8},
 xtick={0,1,2},xticklabels={\shortstack{Layer 1 only\\Historical decay},\shortstack{Layer 1+2\\Pre-window reviews\\correctly excluded},\shortstack{Layer 1+2+3\\Live review stream}},
 xticklabel style={font=\tiny,align=center}]
\addplot[fill=policyblue,draw=none] coordinates {(0,0) (1,0) (2,7)};
\node[font=\tiny\bfseries,align=center,anchor=south] at (axis cs:0,0.05) {0/8\\(structural floor)};
\node[font=\tiny\bfseries,align=center,anchor=south] at (axis cs:1,0.05) {0/8\\(stale signals barred)};
\node[font=\tiny\bfseries,align=center,anchor=south] at (axis cs:2,7.05) {7/8\\(live signals act)};
\end{axis}
\end{tikzpicture}
\caption{Recall on the eight-case set by evidence configuration. History alone: 0/8. History plus pre-window reviews, evaluated correctly: still 0/8 - stale reviews describe conditions an inspector has since checked. Only the live layer moves the number: 7/8.}
\label{fig:3}
\end{figure}
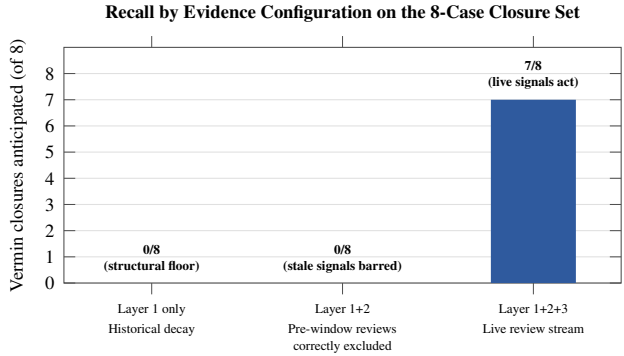

Correcting the filter (Layer 3 must be dated strictly after the most recent inspection) shortened that lead to 209 days and removed one facility's apparent catch entirely (The Counter, whose only review predates its final inspection), changing the headline recall in every configuration. We kept the wrong numbers out of this paper, but we report the mistake itself because it is an easy one: any system that joins an archival review corpus to inspection records will face exactly this boundary, and a signal that predates the last inspection is not evidence about the present. Time-boxed signals such as FINDER's cannot make this error by construction; archival-text systems can.

\section{Layer 3: The Live Trigger}

\subsection{Corroboration, and Its Cost}

With the boundary fixed, we tested the live layer under two configurations, applying the same date filter to positives and negatives alike, and anchoring each comparison facility's window at its second-to-last inspection so that a later inspection cannot retroactively wash out reviews that would have fired in real time. The naive configuration - any single live SAFETY RISK review at full weight, \textit{\ensuremath{\beta} }= 10 chosen by sweep - catches seven of eight closures (The Counter is unreachable in every configuration; Section 7.2) but also flags three of the comparison restaurants, for precision 0.70. The corroboration rule of Section 3 catches exactly the same seven restaurants with exactly the same lead times - including Red Robin, whose single review (a specific report of undercooked chicken and subsequent illness, confidence 0.99, severity 9) clears the severity exception at 0.6\textit{\ensuremath{\times} }weight and fires 126 days before closure - while dropping one false positive: a single confident but low-severity review at a facility with a perfect score, correctly discounted. On this case set, corroboration is free recall-wise and buys precision (0.70 \textit{\ensuremath{\rightarrow} }0.78); its only realized cost is hypothetical, on a facility profile (clean record, one severe review below the exception bar) that this case set happens not to contain.

\begin{table}[H]
\centering
\footnotesize
\begin{tabularx}{\columnwidth}{@{}>{\hsize=1.7\hsize\linewidth=\hsize}Y>{\hsize=.6\hsize\linewidth=\hsize}Y>{\hsize=.8\hsize\linewidth=\hsize}Y>{\hsize=.9\hsize\linewidth=\hsize}Y@{}}
\toprule
\textbf{\textbf{Configuration}} & \textbf{\textbf{Recall (of 8)}} & \textbf{\textbf{Precision}\textbf{\textsuperscript{2}}} & \textbf{\textbf{Mean Lead}} \\
\midrule
Layer 1 only & 0/8 (0.00) & N/A & - \\
Layer 1+3, naive single-review trigger & 7/8 (0.875) & 0.70 & 117 days \\
Layer 1+3, corroboration + severity exception & 7/8 (0.875) & 0.78 & 117 days \\
\bottomrule
\end{tabularx}
\caption*{\textit{Table 3. Case-set results. n = 8 closures (positives), 6--7 comparison facilities (negatives). }\textit{\textsuperscript{2 }}\textit{Precision here measures agreement with closure labels; Section 8 explains why that is not the same thing as correctness. }}
\end{table}

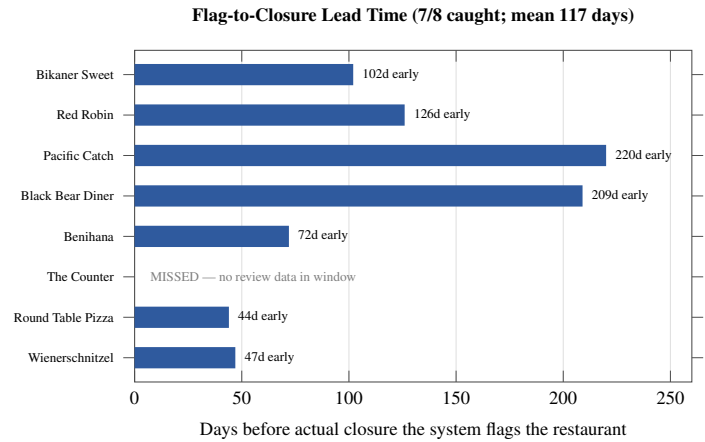
\begin{figure}[H]
\centering
\begin{tikzpicture}
\begin{axis}[paperplot,xbar,bar width=8pt,height=2.35in,
 xmin=0,xmax=260,ymin=-0.6,ymax=7.6,enlarge x limits=false,
 title={Flag-to-Closure Lead Time (7/8 caught; mean 117 days)},
 xlabel={Days before actual closure the system flags the restaurant},
 ytick={0,1,2,3,4,5,6,7},
 yticklabels={Wienerschnitzel,Round Table Pizza,The Counter,Benihana,Black Bear Diner,Pacific Catch,Red Robin,Bikaner Sweet},
 yticklabel style={font=\tiny},xmajorgrids=true,ymajorgrids=false,
 nodes near coords={\pgfmathprintnumber{\pgfplotspointmeta}d early},
 nodes near coords align={horizontal},nodes near coords style={font=\tiny}]
\addplot[fill=policyblue,draw=none] coordinates {(47,0) (44,1) (72,3) (209,4) (220,5) (126,6) (102,7)};
\node[anchor=west,font=\tiny,text=black!50] at (axis cs:3,2) {MISSED --- no review data in window};
\end{axis}
\end{tikzpicture}
\caption{Days between the system's flag and the actual closure, per restaurant. Mean 117 days across the seven catches. The one miss is a data-coverage gap, detailed below.}
\label{fig:4}
\end{figure}

\subsection{The Miss: The Counter}

The Counter's trajectory (80 \textit{\ensuremath{\rightarrow} }82 \textit{\ensuremath{\rightarrow} }72) is a real decline, but a score of 72 cannot lift the decay term to the threshold in the 79 days between its last inspection and closure - Section 6.1's arithmetic again. The only review we hold for this restaurant predates that final inspection by six months and is correctly barred as Layer 2. In other words, the system had nothing to act on: our collection contains no reviews at all for this facility in the December 2025 -- March 2026 window. Whether such reviews exist and were missed by our scrape, or were never written, we do not know; re-collecting that specific window is a named next step, and until then this miss is a statement about our dataset, not about the method.

\subsection{Does the Trigger Actually Fire Promptly?}

The event-driven claim would be hollow if the score only crept over the threshold weeks after a signal arrived. It does not: in four of the seven catches (Pacific Catch, Black Bear Diner, Bikaner Sweet, Round Table Pizza) the threshold is crossed on the same day the qualifying review is posted. The other three require either the severity exception (Red Robin) or a second corroborating review (Benihana, Wienerschnitzel) before firing - which is the corroboration rule behaving as specified, not latency.

\begin{figure}[H]
\centering
\begin{tikzpicture}
\begin{axis}[paperplot,ybar,bar width=28pt,ymin=0,ymax=250,
 title={Four of Seven Catches Fire the Day the Qualifying Review Is Posted},
 ylabel={Days from system flag to closure},
 xtick={0,1,2,3},xticklabels={Pacific Catch,\shortstack{Black Bear\\Diner},\shortstack{Bikaner\\Sweet},\shortstack{Round Table\\Pizza}},
 xticklabel style={font=\tiny,align=center},
 nodes near coords={\pgfmathprintnumber{\pgfplotspointmeta}d}]
\addplot[fill=policyblue,draw=none] coordinates {(0,220) (1,209) (2,102) (3,44)};
\end{axis}
\end{tikzpicture}
\caption{For four of seven catches, one qualifying review moves the facility up the queue the day it is posted.}
\label{fig:5}
\end{figure}
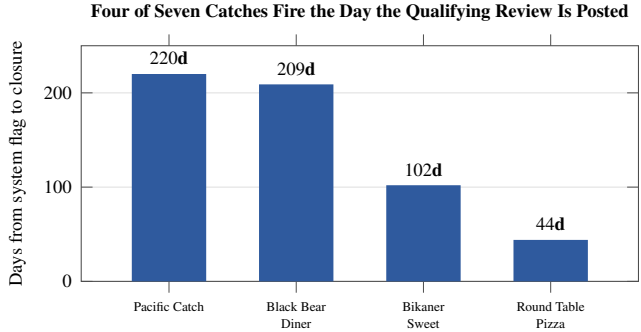

\section{What a False Positive Means Here}

Two of six comparison restaurants are flagged by the final configuration and never closed. We decline to book these as confirmed errors, for a reason that goes to the evaluation design rather than the model. Our only ground-truth label is whether the county eventually closed the facility. A restaurant flagged on genuine, corroborating review evidence that then remediates - cleans up before any inspector visits - produces exactly the same label as a restaurant flagged by mistake. One of our two flagged comparison facilities fits either story equally well: a follow-up inspection occurs shortly after its review signals appear, and its score subsequently rises from 84 to 100. Real problem fixed, or noise? The data cannot say. The retrospective on Chicago's model \papercite{2} wrestles with a structurally similar ambiguity - an observed change in violations found cannot be causally credited to the model without ruling out confounds - and we inherit that limit rather than escape it. Accordingly we report 0.78 as agreement with closure labels, not as a validated precision, and we outline in Section 9 the only design we know of that would break the tie: inspections dispatched on the system's flags, compared against withheld flags.

\section{Limitations and Next Steps}

\begin{itemize}
\item Eight positive cases is a case study. Every number in Table 3 - recall, precision, lead time - is descriptive of this documented set and could move substantially with a handful of additional cases. Scaling to 50--100+ closures across multiple counties is the highest-priority extension.

\item The large-n result (Section 5) and the small-n result (Section 7) evaluate different layers at different scales; they are not two measurements of one deployed system, and only the former is population-level.

\item The 30-review classifier check is the weakest statistical link. The concrete fix: label 100--200 reviews sampled from the full corpus independently of the closure set, with a second annotator and an agreement statistic.

\item The closure-label ambiguity of Section 8 is unresolved and may be unresolvable offline. The decisive experiment is prospective: dispatch some inspections on the system's flags, withhold others, compare findings.

\item The Counter's window (December 2025 -- March 2026) should be re-scraped before the miss is treated as final.

\item Severity values come from a fixed rule table (Appendix B), not from data; \textit{\ensuremath{\beta} }and the corroboration thresholds were tuned on the same eight cases they are evaluated on, and the false-positive check only partially mitigates that circularity.

\item The Layer 1 backtest numbers are pinned to a June 2026 download of the county dataset; a later re-download shifted them by roughly 2--3\% (e.g. 329.7 \textit{\ensuremath{\rightarrow} }323.8 days for M0) as new records arrived. Reproductions should pin the snapshot date.

\item Everything here is offline simulation against historical records. No ingestion pipeline, alerting surface, or inspector workflow was built, and no operational costs were measured.

\end{itemize}
\section{Conclusion}

On a large county dataset, replacing first-come-first-served inspection scheduling with a one parameter decay score finds major violations 69 days sooner on a 30\% budget. On eight documented vermin closures, that same score - and any history-only score at these levels - flags nothing, because the failures occurred at facilities whose paperwork looked fine. Admitting consumer reviews as live evidence, but only reviews postdating the last inspection and only under a corroboration rule with a tested severity exception, recovers seven of the eight with a mean lead of 117 days, usually firing the day the review is posted. Along the way we hit, corrected, and documented a boundary error - stale reviews masquerading as live evidence - that any archival-text system in this space can commit and that inflated one of our own lead times by 100 days before we caught it. And we close on an honest open question: with closure as the only label, a correct flag followed by quiet remediation is indistinguishable from a false alarm, so the next step for this line of work is not more backtesting but a prospective trial. This work provides a case-level existence proof that event-driven inspection triggering survives the move from private telemetry to public data, and contributes a reproducible scoring methodology, a documented failure taxonomy, and a data-hygiene caution for the review-augmented scheduling systems that will follow.

\section*{Appendix A: Parameters}

\begin{table}[H]
\centering
\footnotesize
\begin{tabularx}{\columnwidth}{@{}>{\hsize=.75\hsize\linewidth=\hsize}Y>{\hsize=1.25\hsize\linewidth=\hsize}Y@{}}
\toprule
\textbf{\textbf{Parameter}} & \textbf{\textbf{Value / Description}} \\
\midrule
\ensuremath{\alpha} (score weight) & 1.0 \\
\ensuremath{\lambda} (decay rate) & 0.01 per day \\
\ensuremath{\beta} (live review weight) & 10 (sensitivity sweep, Section 7.1) \\
Scheduling threshold & 70th percentile of risk score across test-set facilities \\
Classifier confidence threshold & 0.75 \\
Single-review discount & 0.3\ensuremath{\times} by default; 0.6\ensuremath{\times} if confidence \ensuremath{\geq} 0.95 and severity \ensuremath{\geq} 8; 1.0\ensuremath{\times} if the last score is < 90 or there are \ensuremath{\geq} 2 qualifying Layer 3 reviews. \\
\bottomrule
\end{tabularx}
\end{table}

\section*{Appendix B: Severity Mapping}

Severity values in live review boost follow a fixed rule table keyed to the classifier's SAFETY RISK sub-category, applied uniformly to every review:

\begin{table}[H]
\centering
\footnotesize
\begin{tabularx}{\columnwidth}{@{}>{\hsize=.75\hsize\linewidth=\hsize}Y>{\hsize=1.25\hsize\linewidth=\hsize}Y@{}}
\toprule
\textbf{\textbf{Sub-category}} & \textbf{\textbf{Assigned Severity}} \\
\midrule
Illness / food poisoning report & 9 \\
Pest sighting (roach, rodent, insect) & 8--9 \\
Raw or undercooked food, contamination & 7 \\
Unsanitary premises (dirty surfaces, sewage odor) & 6--7 \\
Hygiene lapse (bare-hand contact, dirty utensils) & 6 \\
Mold or spoiled food & 6 \\
\bottomrule
\end{tabularx}
\end{table}

\end{document}